\pdfoutput=1
\documentclass[10pt, a4paper]{article}

\usepackage[final]{lrec2026}

\usepackage{natbib}
\usepackage{multibib}
\makeatletter
\def\@mb@citenamelist{cite,citep,citet,citealp,citealt,citepalias,citetalias}
\makeatother

\usepackage{mathtools}
\usepackage{graphicx}
\usepackage{tabularx}
\usepackage{soul}
\usepackage{multirow}
\usepackage{microtype}
\usepackage{latexsym}
\usepackage{tcolorbox}
\usepackage{amsmath}
\usepackage{url}
\usepackage{amssymb}
\usepackage{amsfonts}
\usepackage{graphicx}
\usepackage{tabularx}
\usepackage{multirow}
\usepackage{arydshln}
\usepackage{mathtools,nccmath}
\usepackage{enumitem}
\usepackage{todonotes}
\usepackage{listings}
\usepackage{xcolor}
\usepackage{hyperref}
 \definecolor{darkblue}{rgb}{0, 0, 0.5}
\hypersetup{colorlinks=true, citecolor=darkblue, linkcolor=darkblue, urlcolor=darkblue}

\usepackage{xstring}

\usepackage{color}

\newcommand{\ignore}[1]{}
\title{AccurateRAG: A Framework for Building Accurate Retrieval-Augmented Question-Answering Applications}

\name{Linh The Nguyen$^{1}$, Chi Tran$^{1}$, Dung Ngoc Nguyen$^{1}$, Van-Cuong Pham$^{2}$, \\ {\bf \large Hoang Ngo$^{3}$, Dat Quoc Nguyen$^{1}$}} 

\address{$^{1}$Qualcomm AI Research\textsuperscript{$^{\ast}$}\thanks{$^{\ast}$Qualcomm Vietnam Company Limited. Qualcomm AI Research is an initiative of Qualcomm Technologies, Inc. This work was completed while Van-Cuong Pham and Hoang Ngo were at Qualcomm AI Research.}, $^{2}$University of Oregon, $^{3}$Monash University
\\
\texttt{\small $^{1}$\{linhnt, chitran, dungngoc, datnq\}@qti.qualcomm.com,} \\
\texttt{\small $^{2}$cuongp@uoregon.edu, $^{3}$hoang.ngo@monash.edu}
}

\medskip 

\abstract{
We introduce AccurateRAG---a novel framework for constructing high-performance question-answering applications based on retrieval-augmented generation (RAG). Our framework offers a pipeline for development efficiency with tools for raw dataset processing, fine-tuning data generation, text embedding \& LLM fine-tuning, output evaluation, and building RAG systems locally. Experimental results show that our framework outperforms previous strong baselines and obtains new state-of-the-art question-answering performance on benchmark datasets.
 \\ \newline  
}

\begin{document}

\maketitleabstract

\section{Introduction}

The advent of powerful pre-trained Large Language Models (LLMs) in recent years has unlocked new possibilities for many applications \cite{urlana2024llmsindustriallensdeciphering}. However, relying exclusively on pre-trained models often limits their ability to accurately respond to domain-specific queries or questions about up-to-date information and proprietary knowledge that was not included in their training data. Retrieval-Augmented Generation (RAG) has emerged as a prominent technique to address this limitation \cite{karpukhin-etal-2020-dense,10.5555/3495724.3496517,pmlr-v119-guu20a, gao2024retrievalaugmentedgenerationlargelanguage, yu2024rankrag, zhang2024raft}. 

RAG combines LLMs with external knowledge retrieval mechanisms. Rather than relying only on knowledge implicitly encoded within the LLM's trained parameters, RAG employs an explicit retrieval component to retrieve relevant information from external document collections or knowledge bases. These retrieved contents are then fed into the LLM as additional contexts or references, allowing the model to generate responses grounded in verified external information. 
By incorporating external retrieval, RAG effectively answers queries using customized or specialized datasets \cite{zeng-etal-2024-good,li2024enhancingllmfactualaccuracy}. 

Note that previous RAG papers primarily focus on proposing RAG approaches \cite{10.1145/3637528.3671470,singh2025survey}, not comprehensive frameworks. They do not provide modular components such as a document preprocessor or fine-tuning data generator, which are essential for end-to-end system development. Furthermore, these approaches are independent of the retriever used. 



\begin{figure}[t!]
    \centering
    \includegraphics[width=0.49\textwidth]{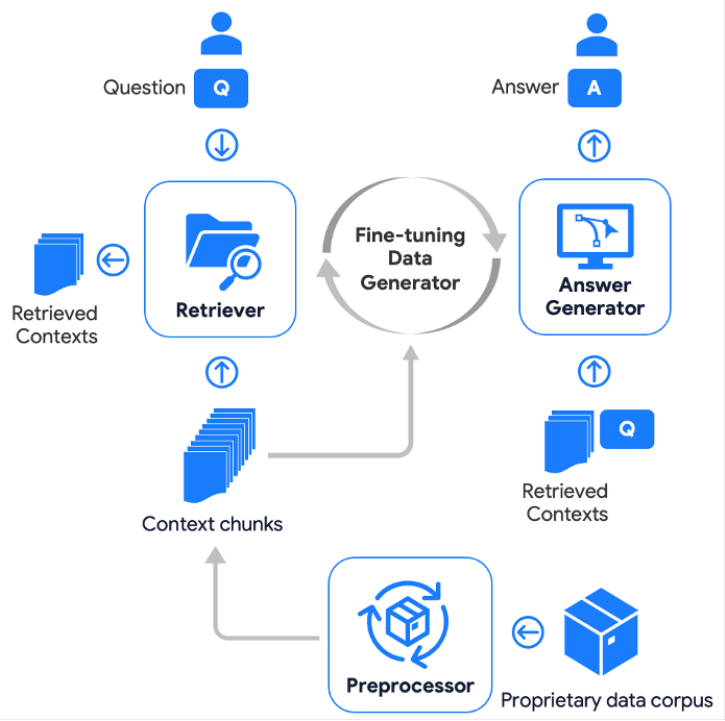}
    \caption{Architecture illustration of our AccurateRAG.}
    \label{fig:AccurateRAG}
\end{figure}

\begin{figure*}[t!]
    \centering
    \includegraphics[width=0.99\textwidth]{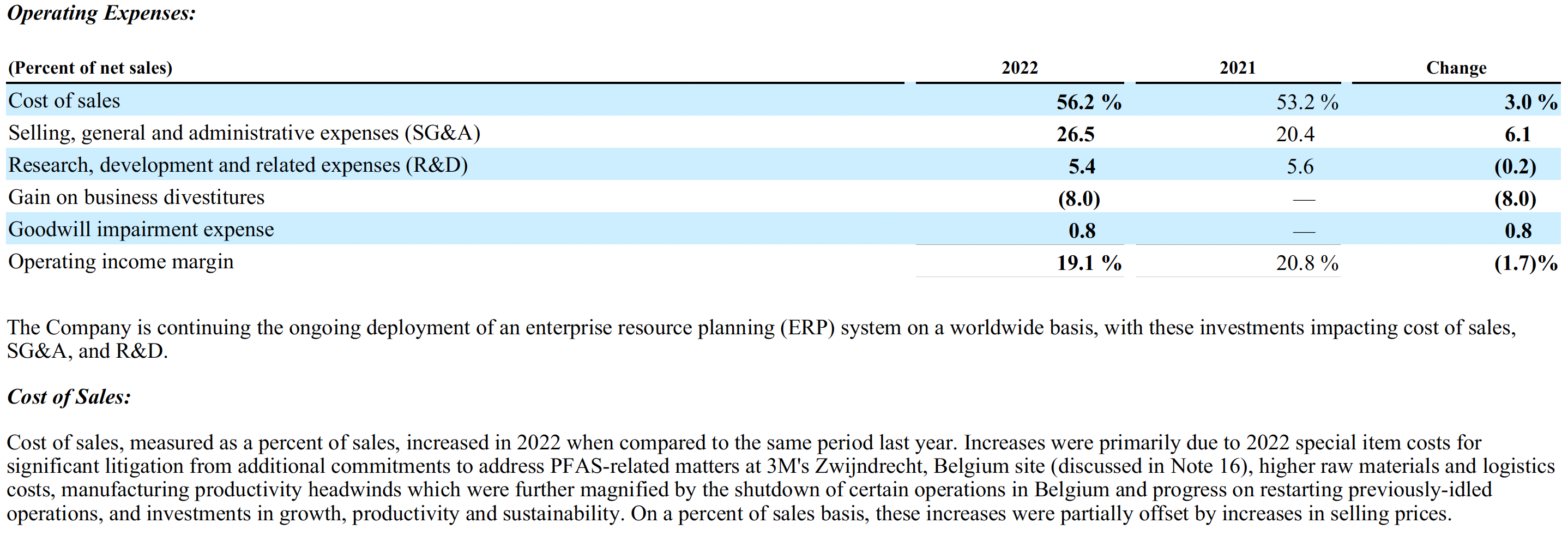}
    \caption{PDF content input.}
    \label{fig:input}
\end{figure*}

\begin{figure*}[t!]
    \centering
    \includegraphics[width=0.99\textwidth]{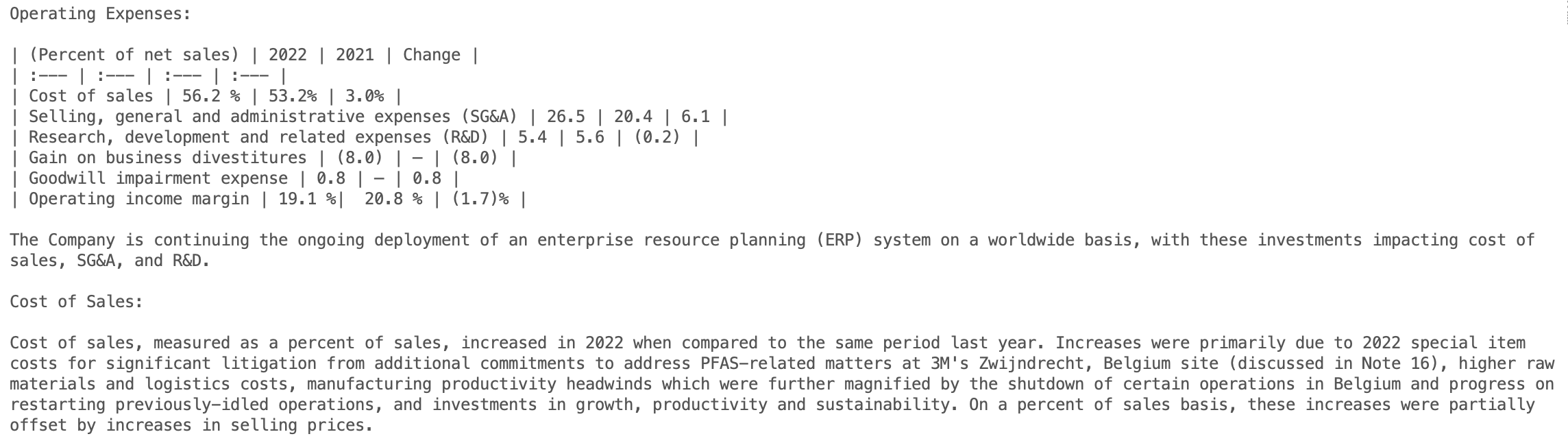}
    \caption{Markdown-formatted text output.}
    \label{fig:output}
\end{figure*}

In light of this, we present AccurateRAG, a novel framework that enables developers to build high-performance RAG-based question-answering applications. 
Our framework offers a pipeline for development efficiency with all the necessary tools to process proprietary datasets (e.g. PDF-to-text conversion with an accurate preservation of structural content), generate question-answering data for model fine-tuning, fine-tune text embedding and LLM models, run evaluations and build RAG systems, all within a local environment. Additionally, it features an intuitive user interface that allows for convenient customization of the system and incorporation of the latest models and data. 
Experimental results show that our AccurateRAG achieves new state-of-the-art performance on standard question-answering benchmark datasets.

\section{AccurateRAG}

Our AccurateRAG framework comprises four components, including: Preprocessor, Fine-tuning Data Generator, Retriever, and Answer Generator, as illustrated in Figure \ref{fig:AccurateRAG}.

\subsection{Preprocessor}

Our Preprocessor is designed to handle data corpora composed of documents in a variety of standard formats, such as PDF and DOCX. Its primary function is to transform each document within the corpus into either plain text, as done in most contemporary systems, or the easy-to-read Markdown format, which allows for a more meticulous preservation of structural content. 

As an illustrative example, a PDF page as depicted in Figure \ref{fig:input} would be converted into a Markdown string as showcased in Figure \ref{fig:output}. In this example, the level of headings and the layout of the table accurately reflect their source, which is more meaningful than reading only the textual content. This demonstrates the Preprocessor's ability to maintain the structure and formatting of the original document during the conversion process. This step is crucial to guarantee that the content is accessible and can be efficiently processed by subsequent components. 
More specifically, we first parse documents into HTML using the Unstructured tool.\footnote{\url{https://github.com/Unstructured-IO/unstructured}} Then, the HTML code for each table element is refined and converted to Markdown using a rule-based method. This approach has one drawback: occasional typos due to Unstructured's imperfect OCR. To mitigate this, we utilize LlamaParse within LlamaIndex, another open-source parsing library that offers higher-quality text but without table structure.\footnote{\url{https://github.com/run-llama/llama_index}} By aligning and combining the outputs of these two libraries, we gain the benefits of both and improve parsing performance immensely.

After the conversion, the Preprocessor divides the parsed data into multiple contiguous context chunks of comparable length to facilitate further analysis. However, we do not require these chunks to have exactly equal lengths. Instead, each context chunk aims to encapsulate a contextual unit. For an intuitive example, a chunk might be shorter to only contain a short subsection, or  longer to cover a lengthy paragraph. This strategy ensures that every retrieved context carries at least one complete, uninterrupted piece of information. We also add to each context chunk a small cut of its preceding and succeeding chunks. This creates a sense of continuity that is especially helpful in multi-hop reasoning use cases.

\subsection{Fine-tuning Data Generator}

The Fine-tuning Data Generator is designed to automatically create synthetic data for the purpose of fine-tuning a text embedding model used in the Retriever and for fine-tuning an LLM model used in the Answer Generator.

In this process, we prompt a pre-trained LLM to generate question-and-answer pairs from each new context chunk. We prompt the LLM to generate multiple simple and complex questions from the given context chunk. Simple questions should be answerable with a single sentence, while complex questions may require more detailed responses, spanning multiple sentences within the context chunk. The questions should cover different aspects of the text to ensure diversity. A further validation step is performed by prompting the LLM to produce the answer for each generated question based on the given context chunk. In this step, questions without answers are removed. This step helps to confirm that the questions make sense in the context of the provided text and that accurate answers are available for them, minimizing the chances of including questions that might seem relevant but are unanswerable or misleading, which can negatively impact the quality of the training data. Our synthetic question-and-answer creation approach thus could expand the training dataset, enabling the model to handle a wider range of queries and improve its overall performance. 

Here, the generated (context, question) pairs are employed for fine-tuning the text embedding model in the Retriever (See the Appendix for examples of generated questions), and the generated (context, question, answer) triplets are utilized for further fine-tuning the LLM in the Answer Generator.

\subsection{Retriever} 

The Retriever consists of three modules: Semantic Search, Conventional Search, and Retrieval Evaluation. 

\subsubsection{Semantic Search}

The semantic search module is to automatically fine-tune a text embedding model based on the (context, question) pairs generated from the Fine-tuning Data Generator, and use the fine-tuned embedding model to find the most relevant contexts for an input question. 

In this approach, we fine-tune a pre-trained BERT-based text embedding model, such as \texttt{bge-large-en-v1.5} \cite{bge_embedding}, using contrastive learning \cite{10.5555/3524938.3525087} with both hard negative examples and in-batch negative examples. To identify a hard negative example for a given question, we utilize the pre-trained BERT-based text embedding model to retrieve the top relevant contexts from all context chunks, explicitly excluding the corresponding positive context from which the question is derived. From these top relevant contexts, we randomly select one context to serve as the hard negative example for the question. To further enhance the fine-tuning process, each batch is constructed with unique positive (context, question) pairs, ensuring that no context is duplicated within the batch. For each positive (context, question) pair in the batch, the contexts from the other pairs within the same batch serve as the in-batch negative examples for the question. This setup encourages the model to distinguish the positive context from both the hard negative and in-batch negatives, improving its ability to create effective text embeddings.

The semantic search module uses the fine-tuned or pre-trained text embedding model to generate embeddings for all questions and contexts. To find the most relevant contexts for a given input question, it calculates the cosine similarity between the question's embedding and the embeddings of the available contexts. The contexts with the highest cosine similarity scores are identified as the most relevant to the input question.

\subsubsection{Conventional Search}

The conventional search module, equipped with the traditional search algorithm BM25 \cite{bm25}, is designed for retrieving the most relevant contexts in response to an input question. BM25 works by ranking documents based on the frequency of the query terms appearing within each document, based on term frequency, inverse document frequency, and document length normalization. When the input question is received, the conventional search module begins by tokenizing and processing the question to identify its key terms. It then compares these terms against an index of pre-processed contexts that have similarly been tokenized and analyzed. The BM25 algorithm evaluates each context based on the frequency of the question's terms, applying its relevance scoring formula to rank contexts according to how well they match the semantic content of the question.

\subsubsection{Retrieval Evaluation}

The retrieval evaluation module assesses various search strategies, including semantic search only, conventional search only, and a hybrid search approach that integrates the top relevant outputs from both semantic and conventional search modules via reciprocal rank fusion \cite{rrf}. This assessment is conducted using a validation set. The module then identifies the most effective search strategy by selecting the one that achieves the highest retrieval score on the validation set. The chosen strategy is subsequently used by the Retriever to find the top most relevant contexts in response to a given question.

\subsection{Answer Generator}

The Answer Generator consists of two modules: Answer Synthesis and Answer Evaluation.

\subsubsection{Answer Synthesis}

The answer synthesis module automatically fine-tunes a pre-trained LLM using ("expanded" context, question, answer) triplets. These triplets are created by combining outputs from the Fine-tuning Data Generator with those from the Retriever. Specifically, for each question in a (context, question, answer) triplet from the Fine-tuning Data Generator, the Retriever is used to identify the top-N-1 most relevant contexts from the entire set of context chunks, explicitly excluding the original context associated with the question. The original context and the top-N-1 relevant contexts for the given question are unified and shuffled to form the "expanded" context. This process results in ("expanded" context, question, answer) triplets that are used for fine-tuning. The answer synthesis utilizes the efficient fine-tuning method LoRA \cite{hu2022lora} to fine-tune the pre-trained LLM on these triplets.

The answer synthesis module employs the fine-tuned LLM (or even the original pre-trained LLM) to generate an answer for a new question based on the concatenation of the top-N relevant contexts provided by the Retriever for the new question.

\begin{figure}[t!]
\begin{tcolorbox}

\noindent You are an expert evaluator. Your task is to determine whether the [Generated Answer] is factually accurate based on the [Query] and the [Ground Truth Answer].

\noindent [Query]: \{query\}

\noindent [Ground Truth Answer]: \{ground truth answer\}

\noindent [Generated Answer]: \{generated answer\}

\noindent \#\# Instructions:

\noindent - Focus only on factual accuracy — ignore style, tone, or completeness.

\noindent - The Generated Answer is accurate if it includes all key facts from the Ground Truth Answer.

\noindent - The Generated Answer is inaccurate if it contradicts or omits key facts from the Ground Truth Answer.

\noindent - Differences in wording, sentence structure, or inclusion of additional context are acceptable.

\noindent Respond with only "TRUE" if the Generated Answer is accurate, and with only "FALSE" if it is inaccurate.
\end{tcolorbox}
\caption{Answer judgment prompt.}
\label{fig:Prompt}
\end{figure}

\begin{figure*}[t!]
    \centering
    \includegraphics[width=1.0\textwidth]{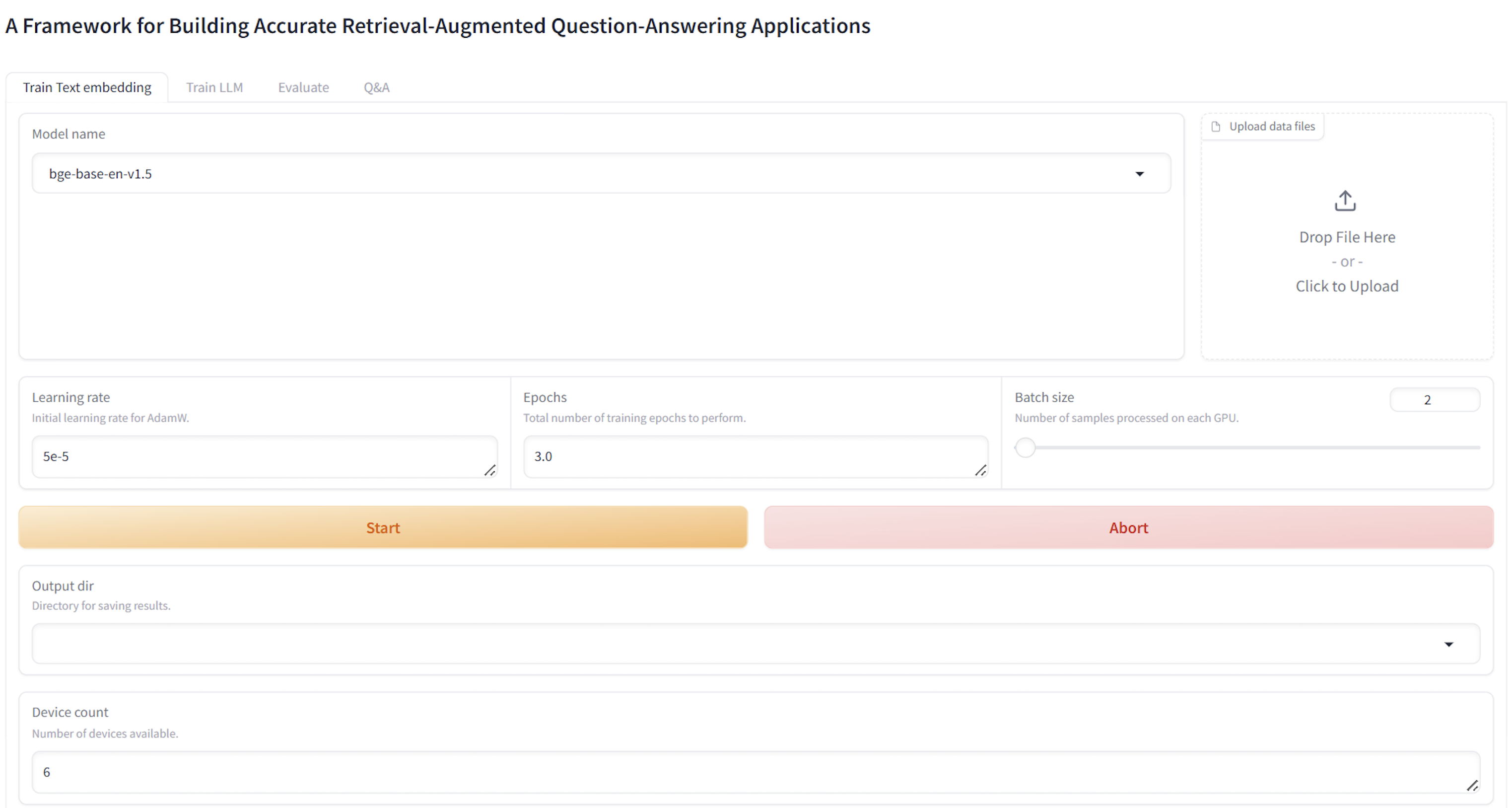}
    \caption{UI for the Preprocessor component and text embedding model fine-tuning in the semantic search module.}
    \label{fig:ftembedding}
\end{figure*}

\begin{figure*}[t!]
    \centering
    \includegraphics[width=1.0\textwidth]{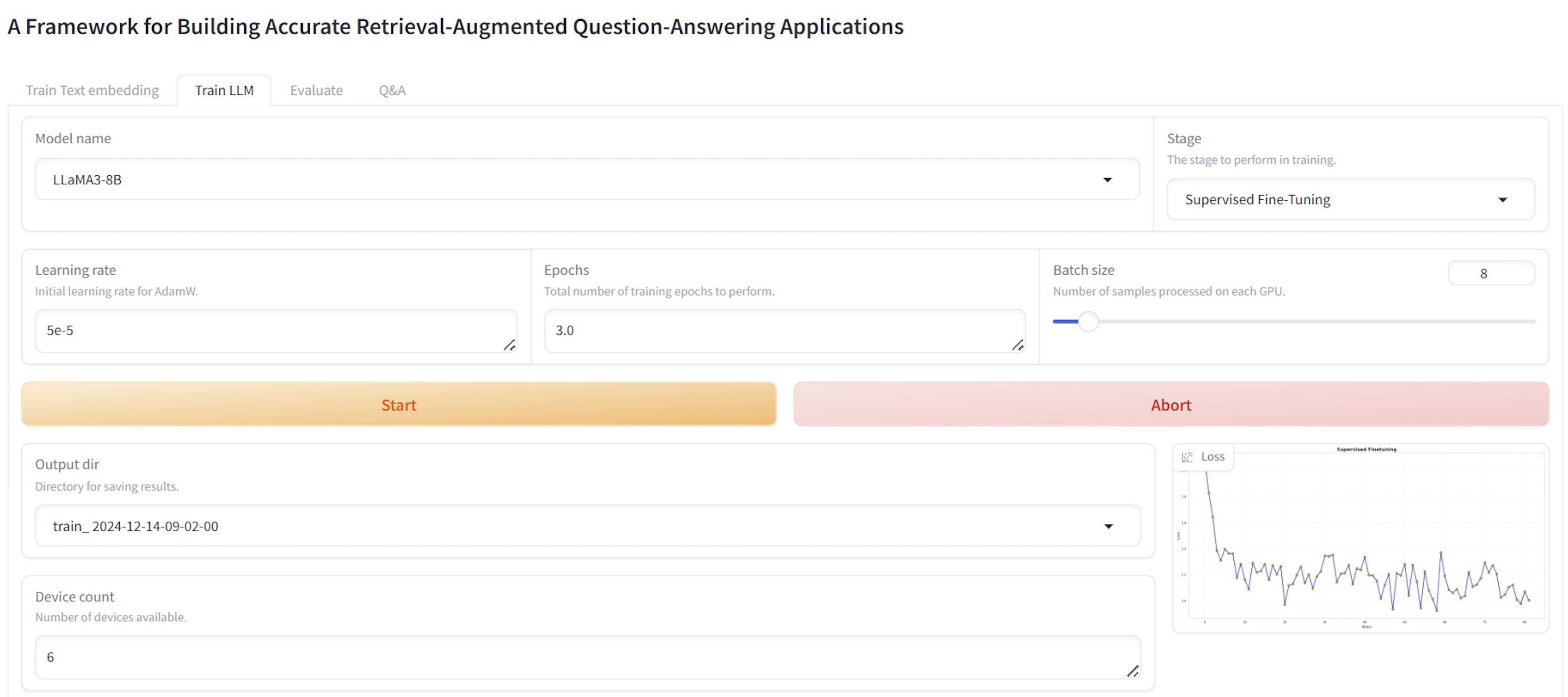}
    \caption{UI for LLM fine-tuning in the answer synthesis module.}
    \label{fig:ftllm}
\end{figure*}

\subsubsection{Answer Evaluation}\label{sssect:Answerevaluation}
The answer evaluation module is designed to use a pre-trained LLM (the \texttt{Llama-3.1-8B-Instruct} model by default) as a judge to evaluate the correctness of the generated answer \cite{zheng2023judging}. Figure \ref{fig:Prompt} shows the prompt used for this module.

\subsection{User Interface}

We also provide a User Interface (UI) to make running AccurateRAG straightforward and efficient, making it more accessible to RAG developers who may not be familiar with command-line operations.

Figure \ref{fig:ftembedding} shows the UI for the Preprocessor component and text embedding model fine-tuning  in the semantic search module, while Figure \ref{fig:ftllm} shows the UI for LLM fine-tuning in the answer synthesis module. The "Evaluate" and "Q\&A" tabs in both Figures \ref{fig:ftembedding} and \ref{fig:ftllm}, though very standard in concept, are designed to support developers in benchmark analysis and interactive QA demonstrations, respectively, ensuring practical usability in real-world scenarios. The remaining modules or components operate in the back-end when triggered. In addition, when a fine-tuning process is initiated, the AdamW optimizer \cite{loshchilov2018decoupled} is employed to facilitate the fine-tuning.

More specifically, as shown in Figure \ref{fig:ftembedding}, developers begin by uploading a private raw data corpus of document files. They then select a pre-trained text embedding model, such as the local folder path of an available BERT-based model or a model identifier from Hugging Face, and specify the fine-tuning hyper-parameters. Upon clicking the "Start" button, the framework first initiates the Preprocessor and subsequently the Fine-tuning Data Generator in the back-end to create context chunks as well as fine-tuning examples. Following this, the semantic search module is executed to fine-tune the text embedding model.

As depicted in Figure \ref{fig:ftllm}, once the text embedding model fine-tuning is complete, developers select a pre-trained LLM and specify the LLM fine-tuning hyperparameters. Upon clicking the "Start" button, the framework first initiates the Retriever, which combines its output with the output from the Fine-tuning Data Generator to create ("expanded" context, question, answer) triplet examples for fine-tuning the LLM, and then executes the LLM fine-tuning process.

\begin{table*}[t]
\centering
\renewcommand{\arraystretch}{1.1}
\resizebox{16cm}{!}{
\begin{tabular}{l|c}
\hline
\textbf{Model} & \textbf{Accuracy (\%)} \\
\hline
\texttt{textembedding-ada-002} (document-level) + \texttt{GPT-4-Turbo} \cite{islam2023financebench} & 19.0 \\
\hdashline
Our \textbf{AccurateRAG}: BGE embedding (w/ FT) + \texttt{GLM-4-9B-Chat} & \textbf{42.0} \\
\hdashline
Our \textbf{AccurateRAG} Preprocessor + BGE embedding (w/o FT) + \texttt{GLM-4-9B-Chat}	& \underline{38.7} \\
\hdashline
\texttt{Unstructured} "fast" Pre-Processing + BGE embedding (w/o FT) + \texttt{GLM-4-9B-Chat} & 	34.7 \\
\hdashline
\texttt{Unstructured} "hi-res" Pre-Processing + BGE embedding (w/o FT) + \texttt{GLM-4-9B-Chat} &	26.7 \\
\hline
\end{tabular}
}
\caption{Question answering results on the FinanceBench test set. "BGE embedding (w/ FT)" denotes the BGE embedding model fine-tuned on generated (context, question) pairs. "BGE embedding (w/o FT)" denotes the original BGE embedding model without further fine-tuning on the generated pairs. \textbf{NOTE  that} {our reported scores are based on manual human verification}.}
\label{tab:financebench}
\end{table*}

\begin{table*}[t!]
\centering
\renewcommand{\arraystretch}{1.1}
\resizebox{16cm}{!}{
\begin{tabular}{l|ccccc}
\hline
\textbf{Model} & \textbf{HotpotQA}	& \textbf{PubMedQA}	& \textbf{HF}	& \textbf{Torch Hub}& \textbf{TF Hub} \\
\hline
RankRAG [w/ FT \texttt{Llama-3-8B}] & 35.30	& 65.0	& N/A	& N/A	& N/A \\
\hdashline
Our \textbf{AccurateRAG} & \textbf{48.71}		& \textbf{82.4}	& 	\textbf{77.21}		& \textbf{93.55}		& \textbf{88.91} \\
\hdashline
RAFT [w/ FT \texttt{Llama-2-7B}]  w/ GPT-4 CoT	& 35.28		& 73.3		& \underline{74.00}		& 84.95		& 86.86  \\
\hdashline
Our \textbf{AccurateRAG} [w/ FT \texttt{Llama-2-7B}]		& \underline{45.71}	& 	\underline{74.6}	& 	68.36	& 	\underline{88.71}		& \underline{88.03} \\
\hdashline
RAFT [w/ FT \texttt{Llama-2-7B}]  w/o GPT-4 CoT	& 25.62		& 68.3	& 	59.07		& 86.56		& 83.21 \\
\hline
\end{tabular}
}
\caption{Question answering results on 5 other test sets. "HF", "TF Hub" and "CoT" abbreviate HuggingFace, Tensorflow Hub and Chain-of-Thought respectively. \textbf{NOTE that} PubMedQA is formulated as a multiple-choice question-answering task, so calculating the accuracy on PubMedQA is straightforward and does not require manual verification. We compute the accuracy on HotpotQA based on the {exact matching} of answer outputs. For the APIBench datasets, following previous work, we employ the standard Abstract Syntax Tree matching evaluation script~\cite{patil2023gorilla}.}
\label{tab:raftbench}
\end{table*}

\section{Evaluation}\label{sec:eval}

\subsection{General Setup}

For all our experiments in this section, we employ two A100 40GB GPUs, running for 3 epochs on training sets with 10\% warm-up steps. For fine-tuning the embedding model, we use a fixed learning rate of 1e-5 and a global batch size of 16. We fine-tune LLMs with LoRA adapters with rank 32, using a fixed learning rate of 5e-5 and a global batch size of 64. 

\subsection{Impact of Preprocessor and Fine-tuning Data Generator}


We evaluate our AccurateRAG on the domain-specific FinanceBench benchmark \cite{islam2023financebench}. The public FinanceBench consists of 150 manually curated question-answer pairs derived from about 80 long financial report PDFs. Available only as a test set, it serves as a typical example that mimics real-world use cases. The benchmark is challenging, as evidenced by the baseline system using the OpenAI's \texttt{ada} embedding model and \texttt{GPT-4-turbo}, which achieves only 19\% accuracy. 

The pre-trained  \texttt{Llama-3.1-8B-Instruct} \cite{grattafiori2024llama3herdmodels} is used as the LLM in the AccurateRAG's Fine-tuning Data Generator component. Our AccurateRAG fine-tunes the BGE text embedding model \texttt{bge-large-en-v1.5} \cite{bge_embedding} using generated data for the semantic search module and uses the pre-trained \texttt{GLM-4-9B-Chat} \cite{glm2024chatglm} as the answer generator. Note that there is no validation set in the public FinanceBench, therefore the AccurateRAG's Retriever component uses the semantic search strategy only. 

As shown in Table \ref{tab:financebench}, AccurateRAG achieves a substantially higher accuracy at 42\%. An ablation study using the original text embedding model without fine-tuning shows a 3\% decrease in accuracy (42\% $\rightarrow$ 38.7\%), demonstrating the effectiveness of our Fine-tuning Data Generator component. Furthermore, replacing our Preprocessor with the well-known Unstructured Pre-Processing Tool for PDF-to-text conversion results in a 4\% accuracy drop (38.7\% $\rightarrow$ 34.7\%),  confirming the superior performance of our Preprocessor.

\subsection{Impact of Model Fine-tuning}

Table \ref{tab:financebench} also presents the effectiveness of fine-tuning the text embedding model with generated data. In this subsection, we further evaluate the effectiveness of fine-tuning both text embedding and LLM models, without generated data.

We evaluate AccurateRAG on 5 standard benchmark datasets, including HotpotQA \cite{yang-etal-2018-hotpotqa}, PubMedQA \cite{jin2019pubmedqa}, and APIBench datasets of HuggingFace, Torch Hub and TensorFlow Hub \cite{patil2023gorilla}, 
that are used in the current state-of-the-art (SOTA) system RAFT  \cite{zhang2024raft}. These datasets already provide training, validation, and test sets, so we do not apply preprocessing or generate additional question-answer pairs. We focus on fine-tuning embedding and LLM models using the provided data. 

For all these benchmarks, we fine-tune the BGE text embedding model \texttt{bge-large-en-v1.5} for semantic search. Once the text embedding fine-tuning process is complete, the  Retriever component determines the retrieval strategy based on the validation sets: it uses the semantic search strategy for both HotpotQA and PubMedQA, and the hybrid search strategy for the APIBench datasets.  For answer synthesis, we first fine-tune different LLMs for different benchmarks: \texttt{Llama-3-8B} for HotpotQA and PubMedQA, and \texttt{CodeGemma1.1-7b-it}~\cite{codegemmateam2024codegemmaopencodemodels} for the APIBench datasets.

We show obtained results in Table 2. Compared to other systems RankRAG \cite{yu2024rankrag} and RAFT, our AccurateRAG obtains notably higher scores than both systems, obtaining new SOTA results. Note that our obtained results with AccurateRAG are based on \texttt{Llama-3-8B} or \texttt{CodeGemma1.1-7b-it}, while the RAFT system uses \texttt{Llama-2-7B} \cite{touvron2023llama2openfoundation} fine-tuned with Chain-of-Thought answers from GPT-4 \cite{openai2024gpt4technicalreport}. 

We further conduct experiments with AccurateRAG based on fine-tuning \texttt{Llama-2-7B} for answer synthesis. In this setting, AccurateRAG achieves higher results than RAFT on 4 out of 5 benchmark datasets. Note that RAFT employs Chain-of-Thought answers from GPT-4. 
When RAFT is not fine-tuned with Chain-of-Thought (under similar settings of fine-tuning \texttt{Llama-2-7B}), AccurateRAG achieves substantially higher scores than RAFT: e.g. AccurateRAG scores about 10\% higher on the HuggingFace dataset (68.36 vs. 59.07) and 20\% higher on the HotpotQA dataset (45.71 vs. 25.62).

\section{Conclusion}

We have presented AccurateRAG---a new framework that provides the necessary tools to help developers build high-performance RAG question-answering applications. AccurateRAG outperforms previous strong baselines, achieving new SOTA results on  question-answering benchmarks. 

\section{Limitations} 

In Section \ref{sssect:Answerevaluation}, using a 8B model as a judge to evaluate the correctness of generated answers might exhibit inadequate confidence calibration, thus producing high-certainty judgments that are incorrect. Future work will integrate calibrated confidence estimation techniques alongside a human-in-the-loop strategy to ensure oversight in low-confidence cases. In addition, fine-tuning the LLM for the answer synthesis module could be further enhanced by employing a foundation LoRA adapter \cite{foundationadapter}.

\section{References}\label{sec:reference}
\bibliographystyle{lrec2026-natbib}
\bibliography{custom}

\newpage 

\section*{Appendix}

\subsection*{Synthesis Data Examples}

\textbf{Given the context:}

The Company continues to make investments in the implementation of new business systems and solutions, including enterprise resource planning, with these investments impacting cost of sales, SG\&A, and R\&D.

Cost of Sales: Cost of sales, measured as a percent of sales, increased in 2023 when compared to 2022. Increases were primarily due to investments in growth, productivity and sustainability; restructuring charges, and carryover impact of higher energy cost inflation partially offset by lower year-on-year net costs for significant litigation to address certain PFAS-related matters at 3M's Zwijndrecht, Belgium site, higher selling prices, spending discipline, sourcing actions and restructuring benefits.

Selling, General and Administrative Expenses: SG\&A, measured as a percent of sales, increased in 2023 when compared to 2022. SG\&A in 2023 was primarily impacted by pre-tax charges of 10.3 billion and 
4.2 billion in the second and third quarters related to the PWS Settlement and the CAE Settlement, respectively (both discussed in Note 18). SG\&A was also impacted by restructuring charges (see Note 5), divestiture costs (related to separating and preparing the Health Care business for spin-off) and continued investment in key growth initiatives. These impacts were partially offset by 2022 net costs for significant litigation to address Combat Arms Earplugs litigation matters (for which a pre-tax charge of approximately \$1.2 billion was reflected in 2022, discussed in Note 18), certain impairment costs related to exiting PFAS manufacturing (see Note 17), costs related to exiting Russia (see Note 17), divestiture-related restructuring charges (see Note 5), restructuring benefits and ongoing general 3M cost management.

Research, Development and Related Expenses: R\&D, measured as a percent of sales, increased in 2023 when compared to 2022. 3M continues to invest in a range of R\&D activities from application development, product and manufacturing support, product development and technology development aimed at disruptive innovations. R\&D was also impacted by restructuring charges.

Gain on Business Divestitures: In 2023, 3M recorded a pre-tax gain of 36 million related to the sale of assets associated with its dental local anesthetic business net of a previous contingent indemnification obligation from a 2020 divestiture. In 2022, 3M recorded a pre-tax gain of \$2.7 billion related to the split-off and combination of its Food Safety business with Neogen Corporation. Refer to Note 3 for further details.

\medskip 
\noindent \textbf{Simple generated questions are as follows:}

\begin{itemize}
    \item How did cost of sales, measured as a percent of sales, change in 2023 compared to 2022?
    \item What was the pre-tax gain recorded in 2022 related to the split-off and combination of the Food Safety business?
\end{itemize}

\medskip 
\noindent \textbf{More complex generated questions are as follows:}

\begin{itemize}
    \item What were the major reasons for the increase in SG\&A expenses in 2023 compared to 2022?
    \item How did R\&D expenses change in 2023 compared to 2022, and what were the contributing factors?
\end{itemize}

\end{document}